# Robustness analysis of Bayesian networks with local convex sets of distributions*

**Fabio Cozman**
Robotics Institute, School of Computer Science, Carnegie Mellon University
fgcozman@cs.cmu.edu, http://www.cs.cmu.edu/~fgcozman

## Abstract

Robust Bayesian inference is the calculation of posterior probability bounds given perturbations in a probabilistic model. This paper focuses on perturbations that can be expressed locally in Bayesian networks through convex sets of distributions. Two approaches for combination of local models are considered. The first approach takes the largest set of joint distributions that is compatible with the local sets of distributions; we show how to reduce this type of robust inference to a linear programming problem. The second approach takes the convex hull of joint distributions generated from the local sets of distributions; we demonstrate how to apply interior-point optimization methods to generate posterior bounds and how to generate approximations that are guaranteed to converge to correct posterior bounds. We also discuss calculation of bounds for expected utilities and variances, and global perturbation models.

## 1 INTRODUCTION

Robust Bayesian inference is the calculation of posterior probability bounds given perturbations in a probabilistic model [3, 22, 38]. This paper presents robust inference algorithms when local perturbations to Bayesian networks are modeled by polytope-like convex sets of distributions.

We consider two ways of defining the combinations of local information in a Bayesian network when convex sets are present.

*This research is partially supported by NASA under Grant NAGW-1175; Fabio Cozman was supported under a scholarship from CNPq, Brazil.

The first approach generates the largest set of joint distributions that satisfies all constraints from local perturbations. We present the first algorithmic analysis of this approach in the context of graphical models. We show how the robust inference problem can be reduced to a fractional programming problem and then solved through standard linear programming techniques.

The second approach takes the convex hull of all combinations of vertices in the local convex sets. We discuss exact algorithms for this problem using the Cano/Cano/Moral (CCM) transform [5]. Due to the complexity of exact algorithms, we develop two classes of approximation algorithms. Firstly, we demonstrate how to use the CCM transform to generate *interior-point* algorithms that converge to robust inferences; the novel idea is to use the CCM transform to reduce robust inference to a formulation that is similar to the problem of learning Bayesian networks. Secondly, we use Lavine's method to reduce the robust inference problem to a particular case of nonlinear programming for which convergence to a global optimizer is assured.

We discuss generalizations of the results to expected utility and variance problems, and present the implementation of local robustness analysis algorithms in the *JavaBayes* system. We also indicate the existence of global perturbation models, in contrast to the local models considered in this paper.

This paper presents several novel algorithms for exact robust inferences; such results promise to open the field of robust Bayesian Statistics to graphical approaches. We conclude by presenting several challenges for future research in this area.

## 2 LOCAL ROBUSTNESS ANALYSIS OF BAYESIAN NETWORKS

In the real world we can rarely meet all the assumptions of a Bayesian model. First, we have to face imperfections in an agent's beliefs, either because the agent had no time, resources, patience, or confidence to provide exact probability values. Second, we may deal with a group of disagreeing experts, each specifying a particular distribution [27]. Third, we may be interested in abstracting away parts of a model and assessing the effects of this abstraction [7, 18]. For example, in the model of Figure 1, an agent may want to assess the impact of the link between variables $A$ and $B$, or the impact of merging variables $C$ and $D$ into a single variable.

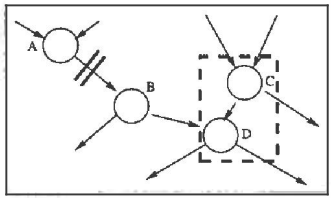

Figure 1: Abstraction in Bayesian networks

Our approach to assessment of robustness is to employ convex sets of distributions to represent perturbations in probabilistic models, both in the prior and conditional distributions. The goal of robustness analysis is to study the impact of such perturbations to posterior values; this is done by analyzing bounds of posterior probabilities.

We use the term *Quasi-Bayesian theory*, as suggested by Giron and Rios [16], to refer to the theory of convex sets of distributions. In this theory there is no commitment to a underlying "true" distribution; a rational decision maker is expected to represent beliefs and preferences through convex sets of distributions which can have more than one element. The basic results of Quasi-Bayesian theory are presented in subsection 2.2. Subsection 2.3 defines two approaches for combination of local information, both of which are studied in this paper.

### 2.1 STANDARD BAYESIAN NETWORKS

We consider a set $\tilde{x}$ of discrete variables; each variable $x_i$ has a finite set of values $\hat{x}_i$ and a set of variables $\text{pa}(x_i)$, the *parents* of $x_i$. A Bayesian network defines a unique joint probability distribution [29]:

$$p(\tilde{x}) = \prod_i p(x_i|\text{pa}(x_i)). \tag{1}$$

We use the abbreviation $p_i$ for $p(x_i|\text{pa}(x_i))$; expression (1) can be written as $p(\tilde{x}) = \prod_i p_i$. Suppose a set of variables is fixed as evidence $e$; $p^e(\cdot)$ is a distribution where variables $e$ are fixed. Our algorithms assume efficient computation of posterior marginals in a Bayesian network [12, 21, 40].

### 2.2 QUASI-BAYESIAN THEORY AND POLYTOPIC CREDAL SETS

Quasi-Bayesian theory uses convex sets of distributions to represent beliefs and to evaluate decisions [16]. Several other theories use similar representations: inner/outer measures [17, 19, 30, 36], lower probability theory [4, 8, 15, 35]), convex Bayesianism [23], Dempster-Shafer theory [34], probability/utility sets [33].

The convex set of distributions maintained by an agent is called the *credal* set, and its existence is postulated on the grounds of axioms about preferences [16]. To simplify terminology, we use the term credal set only when it refers to a set of distributions containing more than one element. Convex sets of conditional distributions are used to represent conditional beliefs. Inference is performed by applying Bayes rule to each distribution in a prior credal set; the posterior credal set is the union of all posterior distributions[1].

We use two well-known results about posterior credal sets in this paper. First, to obtain a posterior credal set, one has to apply Bayes rule only to the vertices of a prior credal set and take the convex hull of the resulting distributions [16]. Second, to obtain maximum and minimum values of posterior probabilities, we must look only at the vertices of the posterior credal sets [37].

Given a convex set $K$ of probability distributions, a probability interval can be created for every event $A$ by defining lower and upper bounds:

$$\underline{p}(A) = \inf_{p \in K} p(A), \qquad \overline{p}(A) = \sup_{p \in K} p(A).$$

In the remainder of this paper, we will refer either to maximization or minimization procedures; lower and upper bounds are closely related through the expression $\underline{p}(A) = 1 - \overline{p}(A^c)$.

Lower and upper expectations for a function $u(x)$ are defined as:

$$\underline{E}[u] = \inf_{p \in K} E_p[u] \qquad \overline{E}[u] = \sup_{p \in K} E_p[u].$$

A credal set always creates lower and upper bounds of probability, but a set of lower and upper bounds of probability does not define a *unique* credal set [37, section 2.7]. The Quasi-Bayesian approach sidesteps this difficulty by taking convex sets as basic entities.

A *polytopic credal set* is the convex hull of a finite number of probability distributions, i.e., it is a polytope in the space of all probability distributions. As we assume that polytopic credal sets are specified over local (presumably small) structures, we assume that representations of polytopic credal sets in terms of vertices and inequalities can be used interchangeably.

---

[1] An introduction to technical aspects of Quasi-Bayesian theory, with a larger list of references, can be found at http://www.cs.cmu.edu/~fgcozman/qBayes.html.

## 2.3 COMBINING LOCAL INFORMATION

Given a Bayesian network, there is a unique way to obtain a joint distribution [29]. This property does not generalize to Quasi-Bayesian models: given a Quasi-Bayesian network, there are several ways to combine the local conditional credal sets into a joint credal set. This paper focuses on two approaches to combination of local credal sets.

The first approach takes the joint credal set as the largest set of distributions that can generate the specified conditionals. This is in many ways the most natural way to represent the joint credal set as it incorporates all possible constraints in the model. The present paper is the first analysis of this method and its algorithmic implications for Bayesian networks.

The second approach multiplies element-wise all local credal sets and considers the convex hull of all resulting joint distributions. The joint credal set is constructed as follows. For each combination of vertices from the local credal sets, construct a joint distribution by multiplying the conditionals together. Now take the convex hull of all these joint distributions as the joint credal set. For example, consider a Quasi-Bayesian network with two credal sets associated with variables $x_1$ and $x_2$. To produce a joint credal set, take the convex hull of the distributions $(p_{1,j} p_{2,k} \prod_{i>2} p_i)$ for all $j$ and $k$. This technique forms the largest joint credal set whose vertices respect the independence relations displayed in the network [37]. The axiomatic underpinnings and algorithmic properties of this method for Bayesian networks have been studied previously [5, 6].

The method of the first approach is referred to as a *natural extension* and the result of the second method is referred to as a *type-1* combination, to use terms proposed by Walley in a similar setting [37, pp. 453, 455]. Note that the natural extension is not identical to a type-1 combination; in most cases a type-1 joint credal set will be smaller than the largest possible credal set given local constraints.

## 3 JOINT CREDAL SETS BY NATURAL EXTENSION

Robust inference in a Quasi-Bayesian network involves the solution of the following minimization for a queried variable $x_q$:

$$\underline{p}(x_q = a|e) = \min \frac{\sum_{x \notin \{x_q,e\}} p^e(\tilde{x})}{\sum_{x \notin e} p^e(\tilde{x})}. \quad (2)$$

The constraints imposed on this minimization are all linear since the local credal sets are polytopes. There are two types of constraints. First, there are linear constraints on the priors:

$$Ap(x_i) \leq b,$$

where $A$ and $b$ are suitable matrices. Second, there are linear constraints on the conditionals:

$$Cp(x_i|\text{pa}(x_i)) \leq d \Rightarrow C\frac{p(x_i, \text{pa}(x_i))}{p(\text{pa}(x_i))} \leq d,$$

which implies that the constraints on the conditionals are:

$$Cp(x_i, \text{pa}(x_i)) - dp(\text{pa}(x_i)) \leq 0.$$

Take the problem above to be embedded in the space of all possible terms in the joint distribution. In this space, the minimization problem is a *linear fractional program*: the minimization of the ratio of two linear functions subject to linear constraints [32]. To write down the program, we must run a cluster-type of Bayesian inference in the network [21], leaving all nodes associated with credal sets in a single cluster. The probability values in this cluster will be the coefficients in the linear fractional program.

The advantage of approaching natural extension from this point of view is that any linear fractional program can be converted to a linear program through a simple transformation [32], for which standard algorithms exist. We are able to produce an exact robust inference through this approach.

The disadvantage of this approach is that the generation of coefficients in the cluster-based network propagation step may be impractical for Quasi-Bayesian networks with a large number of credal sets. The number of terms in the joint distribution is exponential with the number of variables. For a network with $s$ credal sets, each associated with a node with $m$ values, there are $m^s$ variables in the linear program. On the other hand, the number of constraints may be much smaller. Suppose there are $M$ constraints for each node; there are $Ms$ constraints in the linear program. This asymmetry suggests a resort to the dual linear program, in which there will be $Ms$ variables and $m^s$ constraints [1]. Even though the complexity is still the same for exact solutions, now we can use recent results in the field of linear programming [9], which indicate that problems with large numbers of constraints (compared to the number of variables) can be efficiently solved by discarding many redundant constraints. These techniques provide the solution for robust inferences with natural extension.

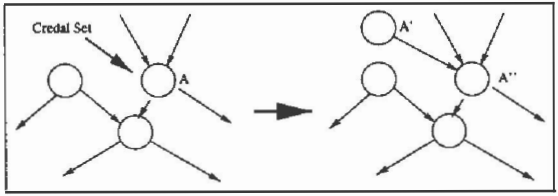

Figure 2: The CCM transformation for variable $A$

## 4 TYPE-1 JOINT CREDAL SETS

This section studies algorithms that determine maximum and minimum values for the posterior distributions generated by a type-1 joint credal set. We begin with the exact solution for this problem using the Cano/Cano/Moral (CCM) transform [5].

### 4.1 EXACT ROBUST INFERENCES

Consider a Quasi-Bayesian network where variables $z_i$ are associated to polytopic credal sets with has vertices $p_{i,j}$. The CCM transformation modifies each variable $z_i$ associated with a credal set (Figure 2):

- Add a new variable $z_i'$ with no parents to the network. The variable $z_i'$ has $z_i$ as its only child. If the credal set for variable $z_i$ has $m_i$ vertices, then $z_i'$ has $m_i$ integer values $\hat{z}_i' = \{1, \ldots, m_i\}$.

- Replace the variable $z_i$ by a new variable $z_i''$ with the same values of $z_i$, all the parents of $z_i$ plus $z_i'$, and the same children of $z_i$. Define the distribution of $z_i''$ to be: $p(z_i''|\text{pa}(z_i'')) = (p_{i,j}(z_i|\text{pa}(z_i))$ when $z_i' = j$).

The variables $z_i'$ are called *transparent* variables [5]. Note that each vertex in the original credal set can be obtained by properly adjusting the transparent variables.

The minimum and maximum values of the posterior credal set can be generated by visiting the vertices of the joint credal set; this can be done by visiting the values of transparent variables.

Suppose one is interested in exact bounds for the posterior $p(a|e)$. One possibility is to cycle over all combinations of transparent variables and use a standard Bayesian inference for each one of them; maxima and minima can be stored as the cycling evolves.

Another exact algorithm comes from trading memory for speed; the idea is to perform a single standard Bayesian inference to obtain the joint distribution $p(a, e, \{z_i'\})$, which includes all transparent variables. Maximization and minimization with respect to the transparent variables produces the required bounds. These algorithms have been hinted in the analysis of the CCM transform [5].

Exact algorithms may be practical in cases where a few credal sets are under consideration, but their complexity grows too fast. If the network has $s$ credal sets, each credal set represented by $m_i$ vertices, there are $\prod_{i=1}^{s} m_i$ independent Bayesian inference runs to be performed. To tackle large problems, we must use approximations.

Cano, Cano and Moral have looked at approximations that treat the selection of transparent variable values as an integer programming problem; they use probabilistic techniques such as simulated annealing and genetic algorithms to handle such problems [5]. We describe two different approaches to this numerical problem. The first approach investigates interior-point methods for its solution (subsection 4.2). The second approach uses Lavine's algorithm to reduce robust inference to signomial programming [1] (subsection 4.3).

### 4.2 INTERIOR-POINT ALGORITHMS

In this subsection we recast the robust inference problem as a parameter estimation problem. Consider a transformed Bayesian network with transparent variables $\{z_i'\}$. Each transparent variable has values $\{1, 2, \ldots, |\hat{z}_i'|\}$. Suppose $z_i'$ is a random variable with distribution $\theta_{ij} = p(z_i' = j)$. Call $\Theta$ the vector of all $\theta_{ij}$.

Suppose $x_q$ is queried; the objective is to find:

$$\overline{p}(x_q = a|e) = \max_{\Theta} \frac{p(x_q = a, e)}{p(e)}$$

Notice that the optimization procedure has to be repeated for each of the values of the queried variable.

To solve the robust inference problem, we must maximize the posterior log-likelihood for $\Theta$:

$$L(\Theta) = \log \frac{p(x_q = a, e)}{p(e)} = \log p(x_q = a, e) - \log p(e).$$

This problem is similar to the problem of learning Bayesian network parameters $\Theta$ given data $e$. We are then lead to propose algorithms for robust inference that are based on the literature of learning Bayesian networks. Several *interior-point* algorithms exist in this category; here we present a few techniques properly adapted for robust inferences.

### 4.2.1 Gradient-based techniques

The gradient of $L(\Theta)$ is obtained by computing, for each $\theta_{ij}$:

$$\frac{\partial L(\Theta)}{\partial \theta_{ij}} = \frac{\partial \log p(x_q = a, e)}{\partial \theta_{ij}} - \frac{\partial \log p(e)}{\partial \theta_{ij}}.$$

This expression (derivation can be found in [10]) is:

$$\frac{\partial L(\Theta)}{\partial \theta_{ij}} = \frac{p(z'_i = j | x_q = a, e)}{\theta_{ij}} - \frac{p(z'_i = j | e)}{\theta_{ij}}, \quad (3)$$

which can be obtained through standard Bayesian network algorithms using local computations. A conjugate gradient descent can be constructed by selecting an initial value for $\Theta$ and, at each step, normalizing the values of $\Theta$ to ensure they represent proper distributions [31].

### 4.2.2 The QEM algorithm

In this subsection we show how the original Expectation-Maximization algorithm [13] can be extended to a Quasi-Bayesian Expectation-Maximization (QEM) algorithm with the same convergence properties. We must maximize the posterior log-likelihood $L(\Theta)$ defined previously. The algorithm begins by assuming that the transparent variables are actual random quantities with distributions specified by $\Theta$. An initial estimate $\Theta^0$ is assumed for $\Theta$.

Suppose we had $i$ sets of complete data for the transformed network, i.e., we had observed $i$ trials for all variables in the network, including the transparent variables. The log-likelihood for this complete data would be $L(\Theta) = \sum_{ijk} l_i(j,k) \log \theta_{ijk}$, where $l_i(j,k)$ indicates the number of data points when the variable $x_i$ is instantiated in its $j$ value with its parents instantiated in their $k$ value.

The first step of the QEM algorithm is to obtain the expected value of the log-likelihood given the evidence and assuming $\Theta^0$ is correct [10]:

$$\begin{aligned} Q(\Theta|\Theta^k) &= E\left[\log\left(p(x_q = a, e)\right) - \log\left(p(e)\right)\right] \\ &= \sum_{ijk} p(x_i, \mathrm{pa}(x_i)|x_q = a, e) \log \theta_{ijk} \\ &\quad - \sum_{ijk} p(x_i, \mathrm{pa}(x_i)|e) \log \theta_{ijk}. \end{aligned}$$

The second step of the QEM algorithm is to maximize $Q(\Theta|\Theta^k)$ for $\Theta$. Only a few terms in the expression for $Q(\Theta|\Theta^k)$ will be free, since only the $\theta_{ij}$ for $z'_i$ are estimated. Collecting these terms we obtain:

$$\sum_{ij} p(z'_i = j|x_q = a, e) \log \theta_{ij} - \sum_{ij} p(z'_i = j|e) \log \theta_{ij}, \quad (4)$$

To perform maximization, use gradient descent with $\Theta^k$ as a starting point and ensure that at the end of the process we have $Q(\Theta^{k+1}|\Theta^k) > Q(\Theta^k|\Theta^k)$. The gradient has essentially the same expression used in the previous subsection, which can be obtained through standard Bayesian network algorithms. Now set $\Theta^{k+1}$ to the maximizing value and go to the next iteration. The following theorem provides the justification for the QEM algorithm (proof can be found in [10]):

**Theorem 1** *The QEM algorithm produces a sequence that converges globally to a local maximum of $L(\Theta)$.*

### 4.2.3 Sampling-based techniques

Once we recast the robust inference problem as the estimation of parameters $\Theta$, we can also assume the parameters $\Theta$ to be assigned uniform priors. In this case, any value of $\Theta$ that maximizes $p(x_q = a, \Theta|e)$ also maximizes $p(x_q = a|e, \Theta)$. We are interested in algorithms that produce such maximizing values of $\Theta$, since $\overline{p}(x_q = a|e) = \max_\Theta p(x_q = a|e, \Theta)$. With this maneuver, we can use Bayesian learning methods to produce robust inferences.

Sampling algorithms for calculation of posterior maxima have been studied in connection with Bayesian inference in general [39]. The reasoning in the previous paragraph demonstrates that they can be applied directly to Quasi-Bayesian inferences as well. Simulated annealing can guide a Gibbs sampler in generating samples of the posterior distribution; the sample with the highest probability defines the maximum [39].

This sampling approach offers a contrast between the interior-point methods advanced here and combinatorial optimization methods that search for the best combination of transparent variable values [5]. In combinatorial approaches, each iteration of the sampling procedure demands a complete cycle of standard Bayesian inference. Instead, by searching in the interior space of distributions, we can use the simulated annealing and Gibbs sampling *simultaneously*; the convergence of this process is a particular benefit of the probabilistic structure of graphical models [39] which is not exploited by purely combinatorial approaches [5].

### 4.3 LAVINE'S BRACKETING ALGORITHM

The previous numerical approaches produced algorithms that converge to optimizers of the posterior distribution, without guarantees about global optimality. In this subsection we sketch an approach to obtain convergence to the global minimum of the posterior distribution. Empirical tests are due to study the practical applicability of this approach.

Lavine's bracketing algorithm is a method to obtain the posterior quantity $\underline{p}(x_q = a) = \min(p(x_q = a, e)/p(e))$. The idea is to settle for deciding whether or not $\underline{p}(x_q = a)$ is larger than a given value $k$. When we obtain this result, we can construct an algorithm by bracketing the interval $[0, 1]$ with $k$. This algorithm is convergent and improves monotonically.

Notice that $\underline{p}(x_q = a) = \min(p(x_q = a, e)/p(e))$ is larger than $k$ if and only if $\min(p(x_q = a, e) - kp(e))$ is larger than zero. The point of Lavine's algorithm is that minimization of the latter quantity may be simpler than minimization of the former, since there are no ratios involved. This is in fact true for type-1 combinations in Quasi-Bayesian networks. Consider the expression that must be minimized:

$$\sum_{x \notin \{x_q, e\}} \prod_i p_i^e - k \sum_{x \notin e} \prod_i p_i^e.$$

This expression is a summation of polynomial terms with arbitrary coefficients subject to linear constraints. This type of problem is termed a *signomial program*, for which there are algorithms that can determine the global minimum [2]. The combination of Lavine's algorithm and signomial programming leads to an algorithm that surely converges to the correct lower bound of the posterior distribution.

## 5 EXPECTED UTILITY AND VARIANCE

This paper has so far concentrated on algorithms for posterior marginals. Most algorithms presented in this paper extend readily to expected utility by simple inclusion of the utility functions [10].

Calculation of lower and upper variances is more complex than expected utility because variances are non-linear functionals of the distributions. We must reduce calculation of variances to an iterative calculation of expected utilities [37, Theorem G2] in order to solve this problem (the method is presented in [10]).

## 6 LOCAL ROBUSTNESS ANALYSIS IN *JavaBayes*

In this section we describe an implementation of local robust analysis for Quasi-Bayesian networks and present an example to illustrate the methods.

Local robust analysis is available in the *JavaBayes* system, a portable and freely distributed inference engine for graphical models. *JavaBayes* is written in Java and can run in any computing platform that supports the Java virtual machine. *JavaBayes* uses standard algorithms to perform calculation of posterior marginals, expectations, maximum a posteriori explanations and maximum a posteriori expectation. Documentation, code and examples for *JavaBayes* can be downloaded from http://www.cs.cmu.edu/~fgcozman/Research/JavaBayes/Home.

| $BatteryPower \rightarrow$ | Good | Poor |
|---|---|---|
| $Lights = Work$ | 0.8 | 0 |
| $Lights = NoLight$ | 0.2 | 1 |
| $BatteryPower \rightarrow$ | Good | Poor |
| $Lights = Work$ | 0.944444 | 0 |
| $Lights = NoLight$ | 0.055555 | 1 |

Table 1: Vertices for the conditional credal set $p(Lights|BatteryPower)$

As an example, consider a troubleshooting problem where the objective is to analyze the state of a car [20], which contains 17 variables and several deterministic and stochastic relationships. Suppose there is some imprecision in the probability values for two variables. First, take the variable *BatteryAge*, which has two values, *Old* and *New*. Suppose this variable is associated with an $\epsilon$-contaminated credal set where $\epsilon = 0.2$ and $p(BatteryAge) = (0.75, 0.25)$ (as detailed in the Appendix). We conclude that this variable is associated with a polytopic credal set with vertices $(0.8, 0.2)$ and $(0.6, 0.4)$. Second, take the binary variable *Lights*, which depends on the binary variable *BatteryPower*. Suppose the expert defines the conditional distribution depicted in Table 1.

This model can be inserted into *JavaBayes* together with arbitrary evidence. For example, if the variable *Starts* is set to *No*, the posterior lower bounds for the binary variable *BatteryPower* are (0.7037, 0.2702) and the posterior upper bounds are (0.7297, 0.2963).

## 7 LOCAL vs. GLOBAL MODELS

This paper investigates local models; credal sets are associated only to marginal or conditional nodes in a network. A different type of Quasi-Bayesian network can be defined through *global* perturbations acting on the whole joint distribution. Several classes of distributions can be used to define such global perturbations (Appendix), but some of them are advantageous from an algorithmic point of view. The $\epsilon$-contaminated, constant density ratio, constant density bounded and total variation classes lead to robust inferences whose complexity is identical to the complexity of standard Bayesian inferences (the algorithms are presented in [11]). Future work will reveal whether such global models are useful for practical robustness analysis.

## 8 CONCLUSION

Robust Bayesian analysis is a field with clear practical relevance; applications of Bayesian networks must determine the relationship between the accuracy of probability values and the accuracy of inferences. Yet research on Bayesian networks has not fully explored this aspect of inference, mostly due to the difficulty of handling probability intervals. Theories of inference that are restricted to linear bounds or belief functions have been plagued by serious mathematical and philosophical difficulties.

The algorithms presented here change dramatically this situation. First, we use Quasi-Bayesian theory, which has a solid, axiomatic foundation, with well-defined analogies for conditioning and decision making. Second, we establish algorithms for general exact and approximate robust inferences. We expect our results to bring robustness analysis to the forefront of tools that are used in a daily basis by Bayesian analysts. However, several issues remain to be addressed. It is necessary to evaluate which algorithms work best with empirical data; a comparison of integer programming methods [5] with interior-point methods is particularly important. Finally, and perhaps most importantly, methods to elicit information about credal sets from experts should be created and evaluated; for example, there must be guidance on how to select between natural extension/type-1 combinations.

## Acknowledgements

We thank Eric Krotkov for substantial support during the research that led to this work, and Lonnie Chrisman for reading an earlier draft and suggesting substantial improvements. We also thank the reviewers, who suggested many improvements and brought important related work to our attention.

## A  Classes of polytopic credal sets

This section demonstrates the generality of our results by placing the most common models of robust Statistics into our framework [10].

**$\epsilon$-contaminated and lower density bounded classes** An $\epsilon$-contaminated class is characterized by a distribution $p(\cdot)$ and a real number $\epsilon \in (0, 1)$:

$$r(x) = (1 - \epsilon)p(x) + \epsilon q(x). \quad (5)$$

An $\epsilon$-contaminated class is the convex hull of the functions $(1 - \epsilon)p(x) + \epsilon \delta_{a_k}(x)$, for all $a_k \in \hat{x}$, where $\hat{x}$ is the set of values of $x$ and $\delta_a(x)$ is 1 if $x = a$ and 0 otherwise. Also, this class is the set of all distributions $p(x)$ so that $p(x) \geq l(x)$ for an arbitrary non-negative measure $l(\cdot)$ [10]; there are approximations (without error bounds) for inferences with this formulation [4].

**Belief function and sub-sigma classes** Consider a discrete variable $x$ with a finite set of values $\hat{x}$. Suppose we impose a probability distribution $m(A)$ into subsets of $\hat{x}$. A *belief function* can be defined from the basic mass assignment as $\text{Bel}(A) = \sum_{B \subset A} m(B)$ [34]; the belief function is the convex set of distributions such that $p(A) \geq \text{Bel}(A)$ for all $A$. To generate the finitely many vertices of the credal set, we must concentrate the non-zero basic mass assignments into each one of their subsets, one at a time.

Consider a variable $x$ with values $\hat{x}$, and a specification of probabilities masses for non-overlapping subsets of $\hat{x}$. This procedure characterizes a sub-class of the belief function class, called a sub-sigma class [3, 24, 28].

**The density bounded class** A density bounded class is the set of all distributions $p(x)$ so that $l(x) \leq p(x) \leq u(x)$, where $l(\cdot)$ and $u(\cdot)$ are arbitrary non-negative measures so that $\sum_x l(x) \leq 1$ and $\sum_x u(x) \geq 1$ [25, 26]. Since finitely many linear inequalities generate this class, it is a polytopic credal set.

**The total variation class** The total variation class is the set of distributions $p(x)$ so that [38]: $|p(A) - r(A)| \leq \epsilon$ for any event $A$, where $r(x)$ is a given probability distribution (a finite number of inequalities is generated this way).

**The density ratio class** A density ratio class consists of distributions $p(A)$ so that for any event $A$ [14]:

$$(l'(A)/l''(B)) \leq (p(A)/p(B)) \leq (l''(A)/l'(B)).$$

where $l'(A)$ and $l''(A)$ are arbitrary positive measures such that $l'(\cdot) \leq l''(\cdot)$. The class is defined by finitely many inequalities, which define intervals of *probability odds*: the ratio between the probability of two events.